\documentclass[letterpaper, 10 pt, conference]{ieeeconf}
\IEEEoverridecommandlockouts
\usepackage{cite}
\usepackage{amsmath,amssymb,amsfonts}
\usepackage{algorithm}
\usepackage{algpseudocode}
\usepackage{graphicx}
\usepackage{comment}
\usepackage{url}
\usepackage{hyperref}
\usepackage{textcomp}
\usepackage{xcolor}
\usepackage[nolist]{acronym}
\usepackage{tikz}
\usepackage{colortbl}
\acrodef{ttc}[TTC]{Time-to-Collision}
\acrodef{dtc}[DTC]{Distance to Collision}
\acrodef{vlm}[VLM]{Vision Language Model}
\acrodef{vqa}[VQA]{Visual Question Answering}
\acrodef{mllm}[MLLM]{Multimodal Large Language Model}
\acrodef{ad}[AD]{Autonomous Driving}
\acrodef{bev}[BEV]{Bird's-eye view}
\acrodef{odd}[ODD]{Operational Design Domain}
\newcommand{\gls}[1]{\ac{#1}}
\newcommand{\glspl}[1]{\acp{#1}}
\newcommand{\acrfull}[1]{\acf{#1}}

\usetikzlibrary{shapes.geometric, arrows, positioning}
\tikzstyle{block} = [rectangle, rounded corners, minimum width=3cm, minimum height=1cm,text centered, draw=black, fill=blue!20]
\tikzstyle{input} = [ellipse, minimum width=2cm, minimum height=1cm, text centered, draw=black, fill=green!20]
\tikzstyle{output} = [ellipse, minimum width=2cm, minimum height=1cm, text centered, draw=black, fill=red!20]
\tikzstyle{arrow} = [thick,->,>=stealth]

\def\BibTeX{{\rm B\kern-.05em{\sc i\kern-.025em b}\kern-.08em
    T\kern-.1667em\lower.7ex\hbox{E}\kern-.125emX}}
\makeatletter
\def\ps@firstpagefooter{%
  \def\@oddhead{}\def\@evenhead{}%
  \def\@oddfoot{\hss\fbox{\parbox{\dimexpr\textwidth-2\fboxsep-2\fboxrule\relax}{\footnotesize This work has been accepted to the IEEE ICRA 2026. Copyright may be transferred without notice, after which this version may no longer be accessible.}}\hss}%
  \def\@evenfoot{}%
}
\makeatother

\begin{document}

\title{NuRisk: A  Visual Question Answering Dataset for Agent-Level Risk Assessment in Autonomous Driving}

\author{Yuan Gao, Mattia Piccinini, Roberto Brusnicki, Yuchen Zhang, Johannes Betz 
\thanks{All authors are with the Professorship of Autonomous Vehicle Systems, TUM School of Engineering and Design, Technical University of Munich, 85748 Garching, Germany; Munich Institute of Robotics and Machine Intelligence (MIRMI)}
}

\maketitle
\thispagestyle{firstpagefooter}
\begin{abstract}
Understanding risk in autonomous driving requires not only perception and prediction, but also high-level reasoning about agent behavior and context. Current \gls{vlm}-based methods primarily ground agents in static images and provide qualitative judgments, lacking the spatio–temporal reasoning needed to capture how risks evolve over time. To address this gap, we propose NuRisk, a comprehensive \gls{vqa} dataset comprising 2.9K scenarios and 1.1M agent-level samples, built on real-world data from nuScenes and Waymo, completed with safety-critical scenarios from the CommonRoad simulator. The dataset provides Bird's-eye view (BEV) based sequential images with quantitative, agent-level risk annotations, enabling spatio–temporal reasoning.  We benchmark well-known \glspl{vlm} across different prompting techniques and find that they fail to perform explicit spatio-temporal reasoning, resulting in a peak accuracy of 33\% at high latency. To address these shortcomings, our fine-tuned 7B VLM agent improves accuracy to 41\% and reduces latency by 75\%, demonstrating explicit spatio-temporal reasoning capabilities that proprietary models lacked. While this represents a significant step forward, the modest accuracy underscores the profound challenge of the task, establishing NuRisk as a critical benchmark for advancing spatio-temporal reasoning in autonomous driving. More information can be found at \url{https://github.com/TUM-AVS/NuRisk}.
\end{abstract}

\section{Introduction}
Autonomous driving has progressed rapidly, with milestones like Waymo’s fully autonomous robotaxi services demonstrating SAE Level 4 capabilities in defined urban environments~\cite{Waymo2018, SAE2021}. These achievements are built on either modular software stacks covering perception, prediction, planning, and control~\cite{pendleton2017perception}, or more recently, end-to-end learning-based approaches~\cite{chen2024end}. However, both paradigms face a fundamental limitation: they struggle to handle the variability of real-world driving, particularly the rare, safety-critical corner cases that fall outside their operational design domains (ODDs)~\cite{Betz2024}. Ensuring trustworthiness requires methods that can reason beyond hand-crafted rules or training data distributions. Recent advances in \acrfull{vlm} offer such potential: by combining scalable, cross-modal knowledge with flexible reasoning, they can improve the coverage of autonomous driving tasks, complementing traditional AD software stacks~\cite{zhou2024vision}.

\begin{figure}[tp]
    \centering
    \includegraphics[width=1\linewidth]{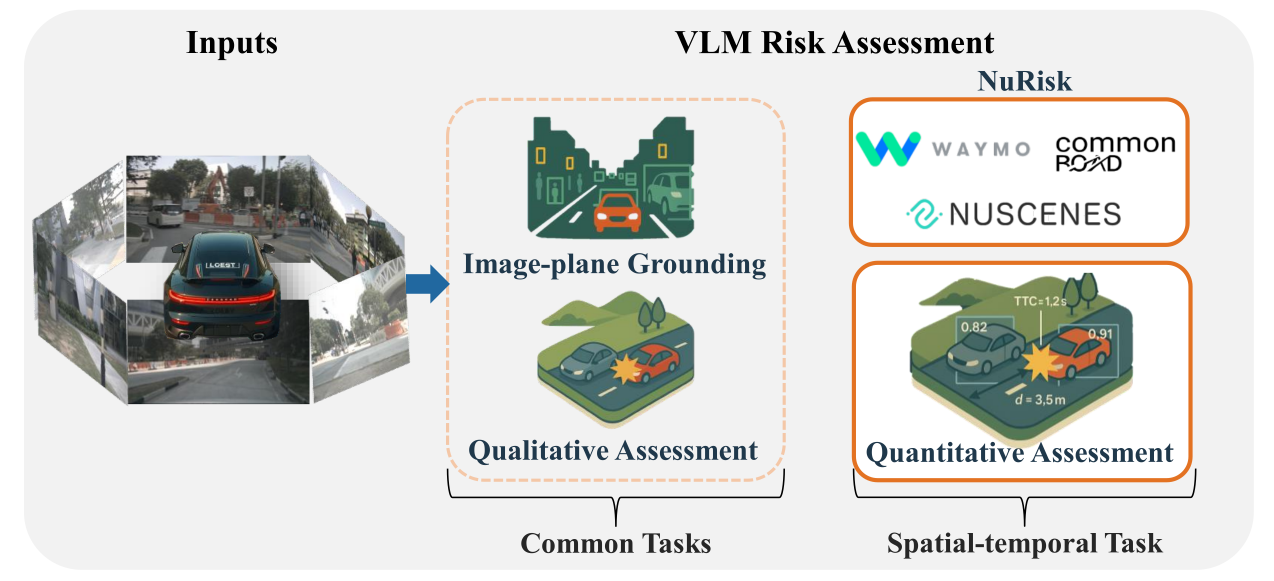}
    \vspace{-4mm}
    \caption{Overview of NuRisk: Existing VLM-based risk assessment is typically limited to (i) image-plane grounding and (ii) qualitative assessment. NuRisk introduces an agent-level quantitative dataset that enables (iii) spatial–temporal quantitative assessment for risk reasoning.}
    \label{fig:concept}
\end{figure}

Despite this potential, the application of \glspl{vlm} to autonomous driving has been focused on scenario risk assessment, such as hazard detection~\cite{zhang2025latte} or qualitative risk assessment~\cite{abu2024using}. While these models can identify a potential hazard, they typically lack the quantitative reasoning required for rigorous safety evaluation. For example, a VLM can identify a vehicle cutting into the ego-car's lane yet offer a qualitative assessment like, "This is a dangerous situation, please drive slowly", which encourages conservative behavior. This raises a question: can pre-trained \glspl{vlm} perform agent-level quantitative risk assessment by leveraging spatio-temporal reasoning to interpret safety metrics, such as the temporal metric \gls{ttc} and the spatial metric \gls{dtc}? Our work, illustrated in Figure~\ref{fig:concept}, confronts this question by proposing a Visual Question Answering (VQA) dataset to evaluate this capability. Such quantitative insights are crucial for downstream motion planners to make informed, precise decisions rather than resorting to overly conservative maneuvers.

\subsection{Related Work}
\subsubsection{VLMs in Autonomous Driving}
Recent studies have explored integrating \glspl{vlm} into autonomous driving systems. Surveys such as~\cite{zhou2024vision, tianlarge} provide comprehensive overviews of the role of \glspl{vlm} in tasks including perception, motion planning, scene understanding, and visual reasoning. In the context of scenario analysis, \glspl{vlm} have shown promising progress. Recently, a comprehensive survey~\cite{gao2025foundation} summarizes four main application areas of \glspl{vlm}-based scenario analysis: \gls{vqa} datasets, scene understanding, benchmarks, and risk assessment.

\subsubsection{VLMs-based Risk Assessment in Autonomous Driving}
Current approaches to \gls{vlm}-based risk assessment can be categorized into two paradigms: prompting-based and fine-tuning-based methods. Prompting-based approaches adapt the reasoning capabilities of large pre-trained models via instructions. GPT-4V has been applied to structured scene representations for risk scoring with natural language justifications~\cite{hwang2024safe}, demonstrating the potential for interpretable risk assessment. Similarly, frameworks like LATTE~\cite{zhang2025latte} and Ronecker et al.~\cite{ronecker2025vision} integrate visual foundation models with contextual prompting for hazard detection and anomaly recognition.

Fine-tuning strategies like LoRA~\cite{hu2022lora} have emerged to address domain-specific requirements and improve robustness by fine-tuning model parameters. Think-Driver~\cite{zhangthink} fine-tunes a \gls{vlm} with chain-of-thought style \gls{vqa} data with multi-view images for hazard reasoning and maneuver risk evaluation. By decomposing risk assessment into sequential reasoning steps, this approach provides more transparent, auditable decision-making than direct predictions. Lee et al.~\cite{lee2025sff} adapt \glspl{vlm} to occlusion-aware BEV representations for uncertainty prediction, while works like INSIGHT~\cite{chen2025insightenhancingautonomousdriving} enhance hazard localization and interpretability. And LKAlert~\cite{wang2025bridging} focuses on failure anticipation by predicting failures in lane-keeping assist systems using multimodal cues. These methods are focused on qualitative agent risk assessment. However, these methods all lack the quantitative labels required for explicit spatio-temporal reasoning, a gap that our NuRisk dataset is precisely designed to fill.

\subsubsection{VQA datasets in Autonomous Driving}
\gls{vqa} datasets for autonomous driving pair visual inputs with natural language queries to evaluate scene understanding across perception, prediction, and planning. Early works enriched perception tasks with BEV maps datasets~\cite{choudhary2024talk2bev}, while later efforts extended to reasoning tasks, including counterfactual reasoning~\cite{wang2024omnidrive} for trajectory generation and video question answering datasets covering perception and prediction~\cite{nie2024reason2drive}. More recent benchmarks push toward multimodal reasoning across the full pipeline, incorporating step-by-step reasoning with images and LiDAR~\cite{ishaq2025drivelmm} or standardized multiple-choice evaluations for \glspl{vlm}~\cite{khalili2025autodrive}. In terms of spatio-temporal reasoning, datasets like NuPlanQA~\cite{ park2025nuplanqa} focus on agents' spatial relations recognition, and TumTraffic-VideoQA~\cite{zhou2025tumtraffic} focus on spatio-temporal grounding in driving scenarios. Additionally, other \gls{vqa} datasets incorporate qualitative agent risk-aware reasoning, such as NuInstruct~\cite{ding2024holistic}, HiLM-D~\cite{ding2023hilm} and DVBench~\cite{zeng2025vision}. However, none of the existing \gls{vqa} datasets systematically evaluate spatial-temporal reasoning for quantitative risk assessment.

\subsection{Critical Summary}
To the best of our knowledge, the existing literature is
limited by at least one of the following aspects:

\begin{enumerate}
\item Insufficient evaluation of spatio-temporal risk assessment: While \glspl{vlm} demonstrate strong performance in static scene risk analysis~\cite{hwang2024safe, zhang2025latte, ronecker2025vision, zhangthink}, their capabilities in reasoning about temporal dynamics, agent trajectories, and evolving risk scenarios remain largely unexplored in safety-critical contexts. Existing approaches predominantly analyze risk at individual time instances rather than understanding how risks develop and propagate over time.

\item Limited scope of existing VQA datasets: Current \gls{vqa} datasets show a clear gap in risk-oriented evaluation. While datasets like NuPlanQA~\cite{park2025nuplanqa} and TumTraffic-VideoQA~\cite{zhou2025tumtraffic} focus on perception and spatio-temporal understanding, and works like NuInstruct~\cite{ding2024holistic} and DVBench~\cite{zeng2025vision} incorporate risk-aware elements, no benchmark systematically evaluates agent-level risk assessment that combines spatio-temporal reasoning with quantitative, safety-critical metrics.

\item Lack of safety-critical scenario coverage: Existing real-world datasets, such as Waymo Open Motion~\cite{ettinger2021large} and nuScenes~\cite{caesar2020nuscenes}, predominantly capture normal driving scenarios, with limited representation of safety-critical situations. This limits the evaluation of \gls{vlm}'s risk reasoning capabilities in high-stakes scenarios, where accurate risk assessment is most crucial to autonomous driving safety.
\end{enumerate}

\subsection{Contribution}
To address the previous limitations, the key contributions of this paper are the following:

\begin{enumerate}
    \item We introduce \textbf{NuRisk}, a \gls{vqa} dataset with 2.9K scenarios and 1.1M agent-level samples for agent-level quantitative risk reasoning, completed with synthetic safety-critical collision scenarios to ensure comprehensive coverage of rare events.
    \item We provide a systematic evaluation of pre-trained off-the-shelf \glspl{vlm} across prompting strategies, highlighting their strengths and limitations in quantitative risk assessment.
    \item We propose a fine-tuning pipeline for open-source \glspl{vlm}, leveraging a parameter-efficient (LoRA) approach to adapt a 7B model for specialized risk assessment tasks, capable of explicit spatio-temporal reasoning, significantly outperforming pre-trained baselines.
\end{enumerate}

\section{Methodology}
\begin{figure*}[htbp]
    \centering
    \includegraphics[width=0.99\linewidth]{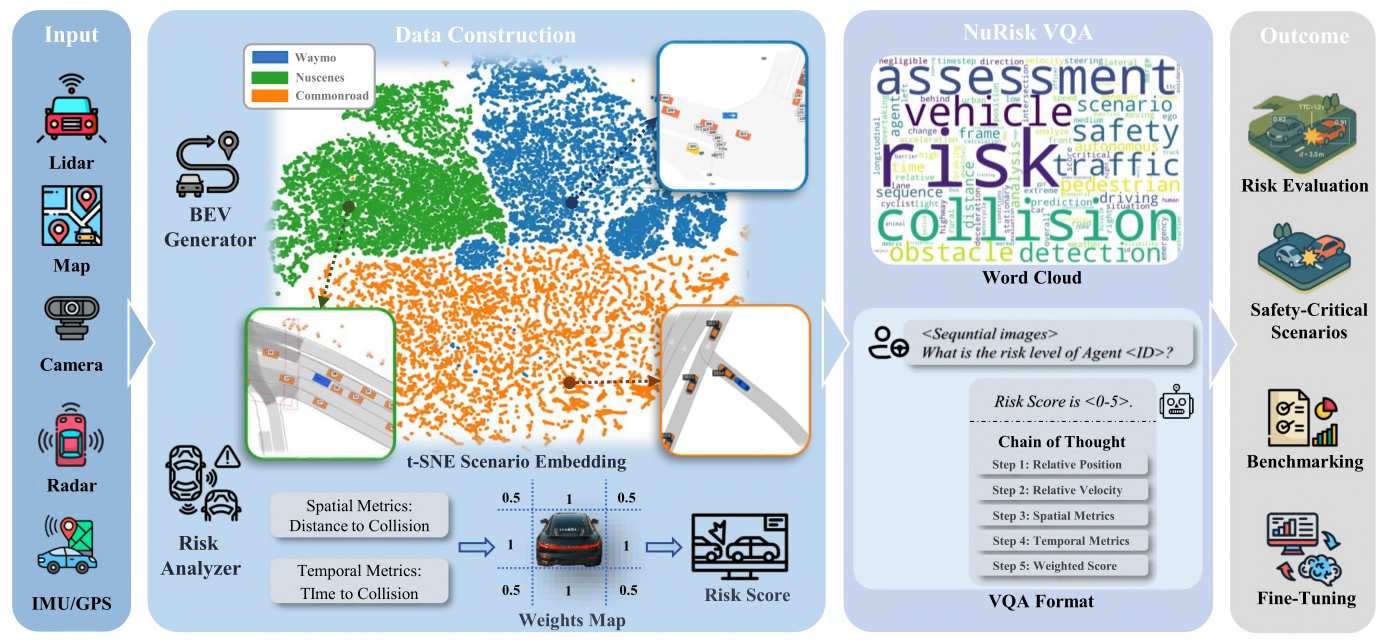}
    \vspace{-4mm}
    \caption{Framework of NuRisk. Multi-modal inputs are processed into BEV scenes and risk metrics to enable conversation-based VQA with chain-of-thought reasoning, supporting risk evaluation, benchmarking, fine-tuning, and safety-critical scenario analysis.}
    \label{fig:framework}
\end{figure*}

This paper introduces NuRisk, a novel \gls{vqa} dataset for agent-level risk assessment in autonomous driving. While high-quality datasets like nuScenes\cite{caesar2020nuscenes} and Waymo Open~\cite{ettinger2021large} are widely used by \glspl{vlm} for their diverse scenarios and multimodal sensor data from Lidar, Camera, Radar, IMU, etc~\cite{gao2025foundation}, a gap exists: none provide ground truth for quantitative, agent-level risk across spatio-temporal scene sequences. To address this, NuRisk utilizes scenarios from these two high-impact datasets and augments them with safety-critical collision scenarios from the CommonRoad simulator~\cite{althoff2017commonroad}. The resulting framework, shown in Figure~\ref{fig:framework}, serves as a base for quantitative risk analysis for individual traffic participants.

\subsection{NuRisk Dataset Construction}
As illustrated in Figure~\ref{fig:dataset} a), we construct NuRisk from 1000 Waymo scenarios, 850 NuScenes scenarios, and 1000 safety-critical scenarios from CommonRoad simulation, where Waymo and NuScenes primarily capture normal driving behavior, and CommonRoad emphasizes safety-critical corner cases. This multi-source composition provides coverage across scenarios characterized by minimum agent risk levels and agent categories. Synthetic scenarios are generated via motion planning in a Commonroad with road networks, traffic signs, and realistic traffic participants. All scenarios are organized into a structured set $S$ that contains driving environment information.
We process these structured scenarios through a multi-stage pipeline shown in Algorithm~\ref{alg:pipeline}:

\textbf{Stage 1: Dataset-Specific Data Extraction.} We extract ego and traffic agent data using dataset-specific methods. For the Waymo Open dataset, we parse TFRecord files to extract ego-vehicle states and all tracked objects, including their trajectories, categories, and dimensional properties. For the nuScenes dataset, we use scene tokens to split the data into individual scenarios and extract ego poses and agent annotations from the structured files, preserving object categories, sizes, and temporal sequences. For CommonRoad scenarios, we obtain ego trajectories from motion planner outputs while extracting other traffic participants from recorded traffic data. Across all datasets, we algorithmically compute velocities and accelerations from position sequences.

\textbf{Stage 2: BEV Image Generation.} 
We adopt \acrfull{bev} representations to isolate spatio-temporal risk reasoning from perception uncertainty. RGB inputs introduce confounding factors such as occlusion and lighting variations, making it difficult to determine whether failures stem from perception limitations or deficiencies in the reasoning ability we aim to evaluate.
We prepare dataset-specific maps and visualization components, then generate BEV images. For nuScenes scenarios, we utilize the map extension API to extract lane boundaries, crosswalks, and road surface information. For Waymo Open datasets, we parse map features from TFRecord files, including lane geometries and traffic infrastructure elements. For CommonRoad, we employ Lanelet2 visualization libraries to render road networks and lane structures. This stage ensures consistent BEV representation across all three data sources while preserving dataset-specific road topology and infrastructure details. For each timestep, the pipeline renders a BEV image focused on a 30-meter radius around the ego vehicle, visualizing all dynamic agents with distinct shapes and colors. Our BEV visualizations, as shown in Figure~\ref{fig:framework}, match standard autonomous driving BEV representations, enabling seamless integration with real-world BEV generation algorithms. 

\textbf{Stage 3: Ground Truth Annotation.}
We compute physics-based risk annotations for each agent in each BEV image. To ensure temporal consistency in the unified dataset, all scenarios are resampled to a uniform 2Hz frequency, following nuScenes’ sampling rate, which is the lowest among the constituent datasets. The pipeline transforms object coordinates into an ego-centric frame, calculates weighted risk scores based on longitudinal and lateral DTC and TTC, and stores the temporally aligned BEV images and risk annotations together in the final dataset $D_{unified}$.

\subsection{VQA Dataset Creation}

After generating the ground truth dataset, we prepare the data for \gls{vlm} training and benchmarking through the following pipeline.

\begin{algorithm}[t]
\small
\caption{BEV Generation and Risk Annotation}
\label{alg:pipeline}
\begin{algorithmic}[1]

\State \textbf{Input:} Scenario Set $S$, Target Frequency $H$, DTC Threshold $\tau_{dtc}$, TTC Threshold $\tau_{ttc}$
\State \textbf{Output:} Unified dataset $D$

\State $D \gets \emptyset$

\For{each scenario $s \in S$}

    \State $(X_e(t), V_e(t)) \gets$ GetEgoTrajectory($s$)
    \State $\{(X_i(t), V_i(t))\} \gets$ GetAgentTrajectories($s$)
    
    \State $T_s \gets$ GetTimestamps($s$)
    \State $T_r \gets$ ResampleIndices($T_s$, $H$)

    \For{each $t \in T_r$}

        \State \Comment{\textbf{--- BEV Generation ---}}

        \State $ego\_state \gets X_e(t)$
        \State $agent\_states \gets \{X_i(t) \mid \text{agent } i \text{ exists at } t\}$
        \State $agent\_states^{ego} \gets$ ToEgo($agent\_states$, $ego\_state$)
        \State $map_t \gets$ GetMapFeatures($s$, $ego\_state$)
        \State $I_t \gets$ Rasterize($map_t$, $agent\_states^{ego}$)

        \State \Comment{\textbf{--- Risk Calculation ---}}

        \State $R_t \gets \emptyset$

        \For{each agent $agent_i$ existing at time $t$}

            \State $\Delta x_i \gets X_i(t) - X_e(t)$
            \State $\Delta v_i \gets V_i(t) - V_e(t)$

            \State $dtc_i \gets \|\Delta x_i\|$
            \State $ttc_i \gets$ ComputeTTC($\Delta x_i$, $\Delta v_i$)

            \State $r^{dtc}_i \gets \text{RiskFromDTC}(\Delta x_i, \tau_{dtc})$
            \State $r^{ttc}_i \gets \text{RiskFromTTC}(\Delta x_i, \Delta v_i, \tau_{ttc})$

            \State $r_i \gets$ CombineRisk($r^{dtc}_i$, $r^{ttc}_i$)

            \State $R^i_t \gets (agent_i,\ dtc_i,\ ttc_i,\ \Delta v_i,\ r^{dtc}_i,\ r^{ttc}_i,\ r_i)$
            \State $R_t \gets R_t \cup \{R^i_t\}$
        \EndFor
        \State $D \gets D \cup \{(I_t,\ R_t)\}$
    \EndFor
\EndFor
\State \Return $D$

\end{algorithmic}
\end{algorithm}

\subsubsection{Preprocessing Pipeline}

The proposed pipeline (Algorithm~\ref{alg:pipeline}) constructs a unified BEV--risk dataset by jointly generating rasterized scene representations and structured risk annotations for all traffic participants.

For each scenario $s \in S$ and each resampled timestep $t$, the algorithm produces a BEV image $I_t$ and a corresponding multi-agent risk set $R_t$. The resulting sequential dataset is defined as
\begin{small}
\begin{equation}
\mathcal{D}_{seq}
=
\left\{
\left(
I_{1:T},
R_{1:T}
\right)
\right\},
\end{equation}
\end{small}
where $I_t$ denotes the BEV representation at timestep $t$, and
\begin{small}
\begin{equation}
R_t
=
\left\{
R_t^{(a)} \mid a \in \mathcal{A}_t
\right\}
\end{equation}
\end{small}
is the set of per-agent risk annotations at time $t$. Each agent-level record is defined as
\begin{small}
\begin{equation}
R_t^{(a)}
=
\left(
\mathrm{id}^{(a)},
\mathrm{dtc}^{(a)},
\mathrm{ttc}^{(a)},
\Delta v^{(a)},
r_{dtc}^{(a)},
r_{ttc}^{(a)},
r^{(a)}
\right),
\end{equation}
\end{small}

For agent-centric modeling, the dataset can be reorganized into per-agent sequences:
\begin{small}
\begin{equation}
\mathcal{D}_{agent}
=
\left\{
\left(
I_{1:T},
R_{1:T}^{(a)}
\right)
\right\}_{a \in \mathcal{A}},
\end{equation}
\end{small}

Subsequently, we perform image optimization to ensure compatibility with \glspl{vlm} input limits, resizing images to meet token constraints while maintaining aspect ratios to preserve visual quality. Finally, we convert the processed data into the following conversation format.
\begin{figure}[t]
    \centering
    \includegraphics[width=0.90\linewidth]{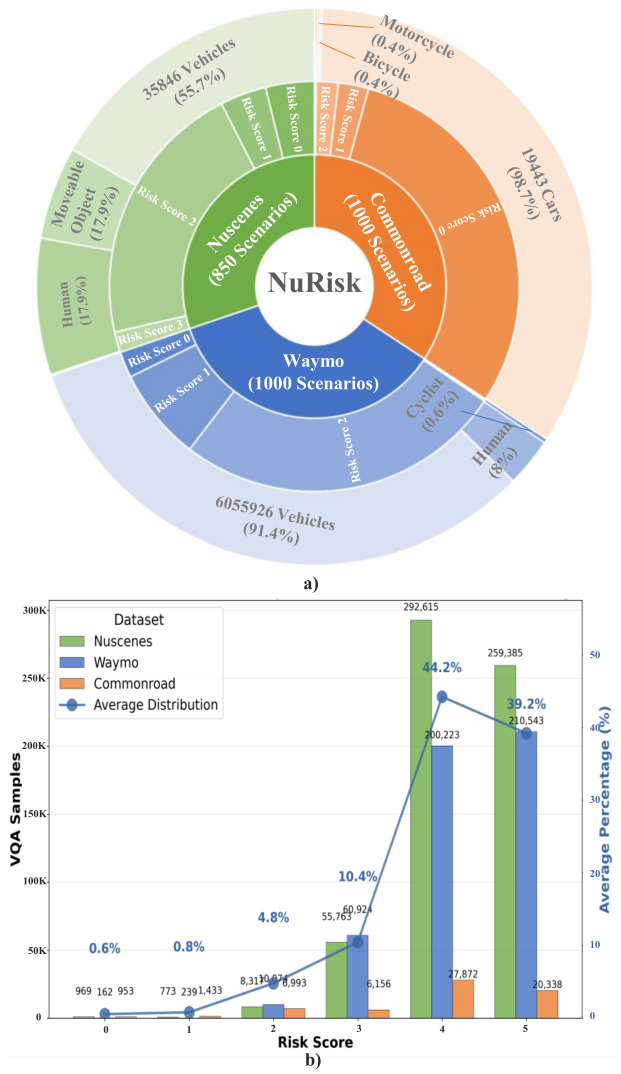}
    \vspace{-4mm}
    \caption{Dataset statistics and risk distribution of NuRisk. Risk scores range from 0 (highest risk/collision) to 5 (lowest risk). \hspace{1em}   (a) Scenario-level composition across data sources, where each scenario’s risk level is defined by the minimum agent risk score within that scenario. Due to its safety-critical design, CommonRoad contains a large proportion of risk-0 scenarios.  \hspace{1em}  (b) Agent-level risk distribution in the final NuRisk dataset. High-risk interactions typically involve only a small subset of agents. Consequently, at the agent level, most agents exhibit lower-risk levels (4–5).}
    \label{fig:dataset}
\end{figure}

\subsubsection{Conversation Format Conversion}
To ensure broad compatibility and leverage a proven instruction-following structure, we adopt the widely used LLaVA conversation format~\cite{liu2023visual}. Each Visual Question Answering sample contains an \textit{image} field specifying the sequential image path and a \textit{conversations} array with alternating \textit{human} and \textit{gpt} messages. \textit{Human} queries include the \texttt{<image>} token and request agent-specific risk analysis, while \textit{gpt} responses provide structured JSON output containing risk assessments with numerical scores, spatio-temporal distance tracking, and chain of thought explanations. This format enables the model to explicitly articulate the reasoning process for predicting agent behavior and assessing risk across temporal sequences, as shown in Figure~\ref{fig:framework}. We also include related word clouds for conversation analysis in the framework visualization.

This format enables direct compatibility with existing \glspl{vlm} training frameworks and ensures consistent, structured outputs for quantitative risk assessment.

\subsubsection{Dataset Quality Assurance} We implement validation throughout the preprocessing pipeline to ensure dataset reliability. Automated quality checks include risk score validation within established ranges, image file integrity verification, conversation pair completeness assessment, JSON structure validation, agent data field verification, and timestep alignment confirmation. Additionally, we conduct human validation of sampled data to verify that sequential images properly display target agents and to cross-validate risk-level assessments against human annotations. The pipeline automatically filters invalid entries and provides detailed processing statistics while maintaining strict quality standards for \gls{vlm} training effectiveness.

The final NuRisk VQA dataset comprises 1.1M agent-level \gls{vqa} samples (617K from nuScenes, 482K from Waymo, 64K from CommonRoad). As shown in Figure~\ref{fig:dataset} b), risk scores are balanced across levels 0-5, ensuring safety-critical coverage for robust \gls{vlm} training,

\subsection{Outcome Analysis}
\subsubsection{Benchmarking for pre-trained VLMs}\label{sec:prompting}
To evaluate off-the-shelf \glspl{vlm} on risk assessment tasks using our proposed NuRisk dataset, we employ a comprehensive evaluation framework that combines zero-shot inference with prompting techniques to adapt pre-trained models for autonomous driving risk analysis.

Our evaluation strategy incorporates three key adaptation techniques: \textit{Contextual Prompting (CP)}, which augments input prompts with task instruction and related driving safety metrics information and collision patterns like safety metrics with explicit thresholds; \textit{Chain-of-Thought reasoning (CoT)}, which enables step-by-step analysis of agent behavior and quantitative risk assessment across temporal sequences; and \textit{In-Context Learning (ICL)}, which leverages selected risk scenario exemplars to establish clear reasoning patterns for understanding different risk driving situations. Results from this evaluation framework, also known as prompting strategies or ablation studies, will be presented in \ref{sec:results}.

\begin{figure}[t]
    \centering
    \includegraphics[width=1\linewidth]{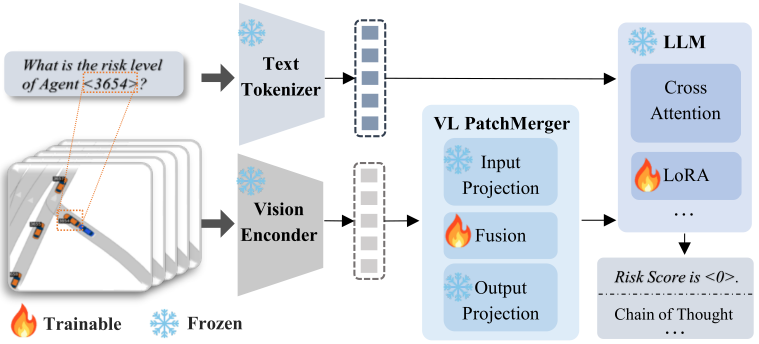}
    \vspace{-8mm}
    \caption{NuRisk VLM Agent Fine-tuning Architecture.}
    \label{fig:finetune}
\end{figure}

\subsubsection{Fine-tuning NuRisk VLM Agent}\label{sec:finetune}
To enhance the spatio-temporal reasoning capability of \glspl{vlm} for autonomous driving agent risk analysis, we fine-tune a \gls{vlm} agent specifically designed for the NuRisk dataset.

We adopt Qwen2.5-VL-7B-Instruct as the base model for its strong multimodal reasoning and moderate parameter count, making it well-suited for parameter-efficient fine-tuning. Its 7B size also remains practical for real-world deployment on edge platforms~\cite{gao2025foundation}. As shown in Figure~\ref{fig:finetune}, the model consists of three main components: (1) a \textit{Vision Encoder}, kept frozen to preserve robust visual feature extraction, (2) a \textit{Language Model}, where base weights are frozen but LoRA adapters are inserted into the attention and feed-forward layers for domain-specific adaptation, and (3) a \textit{Vision Language Merger}, where the input and output projections remain frozen while the Fusion layer is LoRA-adapted to improve cross-modal alignment for risk reasoning. 

We employ causal language modeling for risk assessment conversation completion, optimizing the cross-entropy loss:
\begin{small}
    \begin{equation}
    \mathcal{L} = -\frac{1}{N} \sum_{i=1}^{N} \log p(y_i | x_i, I_i; \theta)
    \end{equation}
\end{small}
where $y_i$, $x_i$, $I_i$, and $\theta$ denote the target response, conversation context, sequential images, and model parameters, respectively. The supervision is provided by structured JSON annotations from the NuRisk VQA dataset. Model selection utilizes validation every 150 training steps, with the optimal checkpoint chosen by $\theta^* = \arg\min_{\theta} \mathcal{L}_{eval}(\theta)$ on held-out validation data, ensuring robust generalization to unseen driving scenarios.
This design enables efficient domain adaptation of large \glspl{vlm}s, strengthening agent-level spatio-temporal risk reasoning without retraining the entire model.

Furthermore, this fine-tuning pipeline is designed to be transferable, as Qwen's architecture shares similarities with the LLaVA framework. The LLaVA conversation format, therefore, provides a standardized training interface, while the LoRA-based parameter-efficient fine-tuning approach enables efficient adaptation to other LLaVA-like VLM architectures without requiring full model retraining.

\section{Results \& Discussion}\label{sec:results}
This section presents our three-stage evaluation. First, we benchmark pre-trained \glspl{vlm} on NuRisk across different prompting techniques (Experiment 1). Second, we incorporate physics-based information to enhance risk assessment (Experiment 2). Finally, we fine-tune a 7B \gls{vlm} on NuRisk to examine whether domain-specific training can close the performance gap (Experiment 3).
\subsection{Experimental Setup}
All experiments were conducted on a workstation equipped with an AMD Ryzen 7 9800X3D CPU, 96 GB of RAM, and an NVIDIA GeForce RTX 5090 GPU. 

\subsubsection{Evaluated Vision Language Models}
Our evaluation includes a suite of leading proprietary and open-source \glspl{vlm}. For proprietary models, we accessed Gemini-2.5-Pro, Gemini-2.5-Flash, GPT-5-Mini, Qwen-VL-Max, and Qwen-VL-Plus via their respective APIs. For open-source models, we conducted local inference using Qwen2.5-VL-7B-Instruct and InternVL3-8B, using the LMDeploy\footnote{\url{https://github.com/InternLM/lmdeploy}} framework.

\subsubsection{Evaluation Metrics}
VLM performance is assessed across three categories. For risk assessment accuracy, we use \textit{Mean Absolute Error (MAE)} to quantify the average magnitude of error in the ordinal risk predictions, \textit{Quadratic Weighted Kappa (QWK)} to evaluate the agreement between predicted and ground-truth risk levels, standard \textit{Accuracy (Acc)}, and \textit{Precision, Recall, F1-Score} for agent risk score analysis. For spatio-temporal reasoning, we measure \textit{Spatial Accuracy} (percentage of longitudinal and lateral predictions within a 0.5-meter tolerance) and \textit{Temporal Accuracy (TTC)} (percentage of longitudinal and lateral predictions within a 0.5-second tolerance). For computational efficiency, we measure the average \textit{Response Time} in seconds. 
\begin{table}[t]
    \centering
    \footnotesize
    \caption{Performance of Pre-Trained VLMs with Vision-Only Input Across Different Prompting Strategies.}
    \begin{tabular}{l@{\hskip 2pt}c@{\hskip 2pt}c@{\hskip 2pt}c@{\hskip 2pt}c}
        \hline
        \textbf{Model + Technique} & \textbf{MAE↓} & \textbf{QWK↑} & \textbf{Acc↑} & \textbf{Time↓} \\
        \hline
        \multicolumn{5}{c}{\textbf{Proprietary Models}} \\
        \hline
        Gemini-2.5-Flash (Baseline)     & 1.91 & 0.49 & 0.15 & 38.29 \\
        \textbf{Baseline + CP}     & \textbf{1.23} & \textbf{0.89} & \textbf{0.33} & \textbf{45.88} \\
        Baseline + CP + CoT    & 1.20 & 0.87 & 0.30 & 96.67 \\
        Baseline + CP + CoT + ICL    & 1.20 & 0.88 & 0.32 & 107.32 \\
        \hline
        Gemini-2.5-Pro (Baseline)       & 2.00 & 0.49 & 0.15 & 40.13 \\
        \textbf{Baseline + CP}       & \textbf{1.25} & \textbf{0.88} & \textbf{0.33} & \textbf{45.24} \\
        Baseline + CP + CoT      & 1.15 & 0.88 & 0.31 & 95.43 \\
        Baseline + CP + CoT + ICL      & 1.22 & 0.86 & 0.30 & 106.55 \\
        \hline
        Qwen-VL-Plus (Baseline)         & 0.63 & 0.55 & 0.13 & 8.81 \\
        Baseline + CP         & 0.58 & 0.62 & 0.17 & 35.81 \\
        Baseline + CP + CoT        & 0.62 & 0.59 & 0.16 & 37.40 \\
        \textbf{Baseline + CP + CoT + ICL}        & \textbf{0.62} & \textbf{0.69} & \textbf{0.22} & \textbf{48.22} \\
        \hline
        Qwen-VL-Max \textbf{(Baseline)}          & \textbf{1.26} & \textbf{0.04} & \textbf{0.22} & \textbf{13.54} \\
        Baseline + CP          & 0.96 & 0.05 & 0.16 & 14.20 \\
        Baseline + CP + CoT         & 1.20 & -0.05 & 0.12 & 26.54 \\
        Baseline + CP + CoT + ICL        & 1.02 & 0.07 & 0.18 & 22.18 \\
        \hline
       GPT-5-Mini \textbf{(Baseline)}           & \textbf{1.16} & \textbf{0.14} & \textbf{0.30} & \textbf{27.67} \\
        Baseline + CP           & 1.23 & 0.07 & 0.25 & 40.49 \\
        Baseline + CP + CoT          & 1.14 & 0.06 & 0.25 & 47.65 \\
        Baseline + CP + CoT + ICL          & 1.24 & 0.05 & 0.25 & 53.34 \\
        \hline
        \multicolumn{5}{c}{\textbf{Open-Source Models}} \\
        \hline
        InternVL3-8B (Baseline)         & 0.62 & 0.56 & 0.14 & 6.55 \\
        Baseline + CP         & 0.55 & 0.70 & 0.20 & 7.15 \\
        \textbf{Baseline + CP + CoT}        & \textbf{0.54} & \textbf{0.70} & \textbf{0.21} & \textbf{11.88} \\
        Baseline + CP + CoT + ICL        & 0.33 & 0.72 & 0.19 & 11.68 \\
        \hline
        Qwen2.5-VL-7B (Baseline)        & 1.87 & 0.51 & 0.14 & 11.98 \\
        Baseline + CP        & 1.88 & 0.46 & 0.12 & 11.60 \\
        Baseline + CP + CoT       & 1.76 & 0.58 & 0.18 & 16.66 \\
        \textbf{Baseline + CP + CoT + ICL}       & \textbf{1.53} & \textbf{0.68} & \textbf{0.22} & \textbf{19.64} \\
        \hline
        \end{tabular}
        \begin{flushleft}
        \scriptsize
        Performance comparison of prompting strategies. Arrows indicate if higher (↑) or lower (↓) values are better. CP: Contextual Prompting, CoT: Chain-of-Thought, ICL: In-Context Learning.
        \end{flushleft}
        \label{tab:prompting_results}
    \end{table}
\subsection{Experiment 1: Benchmarking for pre-trained VLMs}
We conduct our first experiment to evaluate the spatio-temporal reasoning capabilities of pre-trained \glspl{vlm} across different prompting techniques using the NuRisk dataset. This experiment focuses on assessing whether current \glspl{vlm} can perform quantitative risk assessment using mainly image sequences.

Each model is assessed using a baseline model (zero-shot) and with advanced prompting strategies: contextual prompting (CP), chain-of-thought (CoT), and in-context learning (ICL) to establish comprehensive baseline performance across different adaptation techniques, as emphasized in~\ref{sec:prompting}. To provide deeper insights into model capabilities, we assess performance using the metrics defined in our evaluation framework. An analysis examining the effectiveness of different prompting strategies across multiple \glspl{vlm} is presented in Table~\ref{tab:prompting_results}. 

\subsubsection{Proprietary vs. Open-Source Performance}
Proprietary models significantly outperform open-source alternatives, with Gemini variants achieving the highest QWK scores (0.88-0.89) and peak accuracy (0.33). Open-source models reach a maximum QWK of 0.72 (InternVL3-8B) and an accuracy of 0.22 (Qwen2.5-VL-7B).
However, response time is significantly higher for proprietary models than for open-source models, with proprietary models taking 45.88-107.32 s, while open-source models take 6.55-19.64 s.

\subsubsection{Prompting Strategy Effectiveness}
Proprietary models demonstrate a clear preference for informational context over procedural guidance. Contextual prompting can enhance performance by providing task-specific details. Conversely, advanced strategies like chain-of-thought and in-context learning, which dictate reasoning steps, yield negligible improvements while increasing latency. This suggests that these models' internal reasoning mechanisms are not easily augmented by external templates and may even be hindered by them. Furthermore, even for the best proprietary models like Gemini-2.5-Flash or Gemini-2.5-Pro, the moderate accuracy rates ($\leq$0.33) indicate visual information alone is insufficient for reliable autonomous driving risk assessment. While our analysis in Table~\ref{tab:prompting_results} is based on the final risk score, a crucial next step is to examine the full reasoning outputs of these models to better evaluate the nuances of their spatio-temporal reasoning capabilities.

\subsection{Experiment 2: Physics-Enhanced Input Configuration Analysis}
To address the performance limitations observed in Experiment 1, we conduct an evaluation incorporating physics-based information like position, velocity, and acceleration from the ego vehicle and nearby vehicles into the input modalities. From the paper~\cite{gao2025words}, LLMs with physical information from sensor data can already reach more than 0.8 accuracy for the risk assessment task with a hand-crafted prompt. 
We evaluate the performance of both proprietary and open-source models across different input configurations, including single-image and multi-image sequences. We also assess their performance across various prompting strategies, including contextual, chain-of-thought, and in-context prompting. 

\subsubsection{Impact of Input Modality}
As shown in Table~\ref{tab:modality_results}, incorporating textual physics information yields substantial performance gains, particularly for proprietary models. Gemini-2.5-Flash, for instance, achieves a peak accuracy of 0.92 and an MAE of 0.19, significantly outperforming both its vision-only counterpart in Table~\ref{tab:prompting_results} and the text-only Gemini-1.5-Pro baseline from the paper~\cite{gao2025words}. In contrast, open-source models fail to realize similar improvements. This is largely due to their limited context windows, which cannot effectively process token-intensive textual data. For InternVL3-8B, this results in a dramatic increase in response time from 1.36 s to over 32 s.
\begin{table}[t]
    \centering
    \footnotesize
    \caption{Performance Analysis of VLMs with Physics-Enhanced and Sequential-Image Inputs.}
    \begin{tabular}{l@{\hskip 2pt}c@{\hskip 2pt}c@{\hskip 2pt}c@{\hskip 2pt}c}
    \hline
    \textbf{Model + Technique + Modality} & \textbf{MAE↓} & \textbf{QWK↑} & \textbf{Acc↑} & \textbf{Time↓} \\
    \hline
    \multicolumn{5}{c}{\textbf{Proprietary Models}} \\
    \hline
    Gemini-1.5-Pro~\cite{gao2025words}   (Text) & - & - & 0.83 & 25.00 \\
    \hline
    Gemini-2.5-Flash (Baseline) (Single) & 1.08 & 0.44 & 0.11 & 7.66 \\
    Baseline + CP (Single+Text) & 0.21 & 0.83 & 0.91 & 24.51 \\
    Baseline + CP + CoT (Single+Text) & 0.20 & 0.84 & 0.91 & 32.63 \\
    Baseline + CP + CoT + ICL (Single+Text) & 0.22 & 0.82 & 0.90 & 31.57 \\
    Gemini-2.5-Flash (Baseline) (Multi)  & 1.91 & 0.49 & 0.15 & 38.29 \\
    Baseline + CP (Multi+Text) & 0.20 & 0.84 & 0.92 & 26.76 \\
    Baseline + CP + CoT (Multi+Text) & 0.20 & 0.84 & 0.92 & 34.72 \\
    \textbf{Baseline + CP + CoT + ICL (Multi+Text)} & \textbf{0.19} & \textbf{0.85} & \textbf{0.92} & \textbf{40.55} \\
    \hline
    \multicolumn{5}{c}{\textbf{Open-Source Models}} \\
    \hline
    InternVL3-8B (Baseline) (Single) & 1.65 & 0.14 & 0.13 & 1.36 \\
    Baseline + CP (Single+Text) & 1.55 & 0.20 & 0.20 & 1.54 \\
    \textbf{Baseline + CP + CoT (Single+Text)} & \textbf{0.95} & \textbf{0.44} & \textbf{0.39} & \textbf{21.14} \\
    Baseline + CP + CoT + ICL (Single+Text) & 1.23 & 0.39 & 0.30 & 32.84 \\
    InternVL3-8B (Baseline)  (Multi)    & 0.62 & 0.56 & 0.14 & 6.55 \\
    Baseline + CP (Multi+Text) & 1.71 & 0.18 & 0.16 & 2.01 \\
    Baseline + CP + CoT (Multi+Text) & 1.13 & 0.30 & 0.33 & 23.17 \\
    Baseline + CP + CoT + ICL (Multi+Text) & 1.18 & 0.40 & 0.38 & 33.18 \\
    \hline
    Qwen2.5-VL-7B (Baseline) (Single) & 1.95 & 0.12 & 0.12 & 2.63 \\
    Baseline+ CP (Single+Text) & 1.70 & 0.18 & 0.18 & 2.96 \\
    Baseline + CP + CoT (Single+Text) & 1.71 & 0.18 & 0.18 & 13.02 \\
    Baseline + CP + CoT + ICL (Single+Text) & 1.43 & 0.27 & 0.18 & 44.91 \\
    Qwen2.5-VL-7B (Baseline) (Multi)      & 1.87 & 0.51 & 0.14 & 11.98 \\
    Baseline + CP (Multi+Text) & 1.78 & 0.16 & 0.16 & 6.50 \\
    \textbf{Baseline + CP + CoT (Multi+Text)} & \textbf{1.55} & \textbf{0.20} & \textbf{0.25} & \textbf{22.95} \\
    Baseline + CP + CoT + ICL (Multi+Text) & 1.58 & 0.20 & 0.23 & 33.18 \\
    \hline
    \end{tabular}
    \begin{flushleft}
    \scriptsize
    Performance with physics-enhanced inputs. CP: Contextual Prompting, CoT: Chain-of-Thought, ICL: In-Context Learning. Multi/Single refers to Sequential/single-image input.
    \end{flushleft}
    \label{tab:modality_results}
    \end{table}
The addition of sequential-image data provides a more modest benefit. While Gemini-2.5-Flash sees a slight accuracy increase from 0.91 to 0.92, open-source models derive no clear advantage, as the increased token load from multiple images further strains their processing capabilities and increases latency. This suggests that for risk assessment at the current timestep, explicit physics data is the most critical input, with sequential visual context offering only marginal gains.

\subsubsection{Impact of Prompting Strategy}
From the ablation study, we reach the same conclusion as shown in Table~\ref{tab:modality_results}: for pre-trained reasoning models, performance with contextual prompting is better than zero-shot performance. However, other advanced prompting strategies do not improve performance well while substantially increasing response time. This occurs because the proposed prompting strategies do not enhance the internal reasoning of models that already have access to physical information and primarily rely on visual-spatio-temporal capabilities for task reasoning. So \glspl{mllm} look promising for the risk assessment task, since they can reason about the physical information from the other sensors and the visual information.

\begin{figure}[t]
    \centering
    \includegraphics[width=1\linewidth]{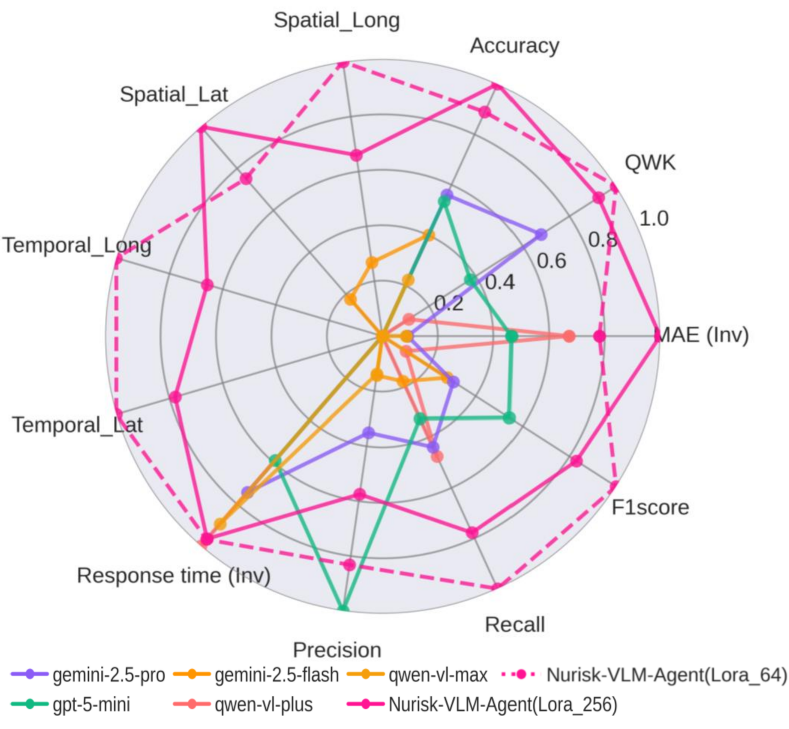}
    \vspace{-6mm}
     \caption{Performance comparison between the best proprietary \glspl{vlm} and our two fine-tuned NuRisk VLM Agent configurations, with the inverse scale of the response time and MAE to make it more interpretable.}
      \label{fig:experiment3}
\end{figure}
\subsection{Experiment 3: Fine-tuning the NuRisk VLM Agent}
Experiments 1 and 2 revealed a critical gap: while proprietary \glspl{vlm} like gemini-2.5-pro with 0.33 accuracy cannot achieve reasonable quantitative risk assessment accuracy and operate as high-latency "black boxes."  To address these shortcomings, we developed the NuRisk VLM Agent by fine-tuning a Qwen2.5-VL-7B-Instruct as introduced in Section~\ref{sec:finetune}. We fine-tuned the agent on 50,000 samples from the NuRisk dataset for 2 epochs, creating 2 variants with LoRA ranks of 64 and 256.

The comparative performance is visualized in Figure~\ref{fig:experiment3}, which is normalized to our NuRisk VLM Agent, and we also invert the scale of the response time and MAE to make it more interpretable. Our fine-tuned agents significantly outperform all proprietary models across nearly every metric, demonstrating the effectiveness of specialized training.

Our fine-tuned agents demonstrate a superior balance of accuracy and efficiency. The LoRA-256 agent achieves a peak accuracy of 41.1\% (MAE: 1.01, QWK: 0.28), representing a significant improvement over its pre-trained baseline accuracy of 14\% and outperforming the best proprietary model (Gemini-2.5-Pro at 33\%). Furthermore, it maintains an average response time of 10.2 seconds, which is 4 times faster than the leading proprietary model, making it more viable for real-world applications. 

The most critical distinction lies in spatio-temporal reasoning. Proprietary models completely fail in this domain, even with the provided collision patterns in the prompt. In contrast, our NuRisk LoRA-256 agent exhibits strong performance, achieving longitudinal spatial accuracies of 34.1\% and lateral spatial accuracies of 26.0\%, and temporal longitudinal accuracies of 27.0\% and lateral accuracies of 26.4\%. This indicates it has learned the underlying causal relationship between vehicle dynamics and risk, proving our agents are not merely classifying risk but are reasoning about spatio-temporal information.
The two fine-tuning configurations offer distinct advantages. The LoRA-256 agent is optimized for maximum classification accuracy. The LoRA-64 agent, while slightly less accurate, provides a more balanced performance profile, achieving the highest QWK score (0.304) and showing stronger performance on temporal reasoning metrics. Both configurations confirm the effectiveness of domain-specific fine-tuning.

\section{Conclusion and Future Work}
We introduced a framework for advancing spatio-temporal VLM reasoning in autonomous driving. We presented the NuRisk dataset with 2.9K scenarios and 1.1M agent-level samples, a diverse collection from nuScenes, Waymo, and CommonRoad. Based on the NuRisk dataset, we showed that even top proprietary models like Gemini-2.5-Pro (33\% accuracy) fail at spatio-temporal reasoning (scoring zero on all spatio-temporal metrics). We address this gap by showing that physics-enhanced inputs can boost accuracy to 92\%. Building on this, we introduced the NuRisk VLM Agent, fine-tuned on a small dataset and achieving 41.1\% accuracy while running four times faster than the best proprietary model. Crucially, our agent is the only model to demonstrate the causal spatio-temporal reasoning necessary for true quantitative risk understanding.
For future work, we plan to expand the NuRisk dataset with more challenging scenarios and further optimize the NuRisk VLM Agent for real-world deployment by lightweight \gls{vlm} models with low latency.
\section*{ACKNOWLEDGMENT}
The manuscript was initially drafted by the authors, with AI tools used to improve grammar and clarity.

\bibliographystyle{IEEEtran}
\bibliography{literature.bib}

\begin{acronym}
\acro{AVs}{autonomous vehicles}
\acro{RL}{reinforcement learning}
\end{acronym}


\end{document}